\documentclass[letterpaper, 10 pt, conference]{ieeeconf}  

\IEEEoverridecommandlockouts                              

\usepackage{amsmath}
\usepackage{amssymb}
\usepackage{dirtytalk}
\makeatletter
\let\NAT@parse\undefined
\makeatother
\usepackage{graphicx}
\usepackage{hyperref}  

\title{\LARGE \bf
A Probabilistic Relaxation of the \\ Two-Stage Object Pose Estimation Paradigm
}

\author{Onur Beker 
\thanks{O. Beker is with the School of Computer and Communication Sciences, Ecole Polytechnique Federale de Lausanne (EPFL), CH-1015 Lausanne, Switzerland (email:
        {\tt\small onur.beker@epfl.ch}). \newline \indent \textit{* This work was presented as an extended abstract in the Workshop on Safe \& Robust Learning for Perception-based Planning and Control, as part of the American Control Conference (ACC) 2023.}}%
}

\begin{document}

\maketitle
\thispagestyle{empty}
\pagestyle{empty}

\begin{abstract}
Existing object pose estimation methods commonly require a one-to-one point matching step that forces them to be separated into two consecutive stages: visual correspondence detection (e.g., by matching feature descriptors as part of a perception front-end) followed by geometric alignment (e.g., by optimizing a robust estimation objective for pointcloud registration or perspective-n-point). Instead, we propose a matching-free probabilistic formulation with two main benefits: i) it enables unified and concurrent optimization of both visual correspondence and geometric alignment, and ii) it can represent different plausible modes of the entire distribution of likely poses. This in turn allows for a more graceful treatment of geometric perception scenarios where establishing one-to-one matches between points is conceptually ill-defined, such as textureless, symmetrical and/or occluded objects and scenes where the correct pose is uncertain or there are multiple equally valid solutions.

\end{abstract}


\section{Mathematical Framework}

\begin{figure*}[th]
  \centering
  \includegraphics[width=\textwidth]{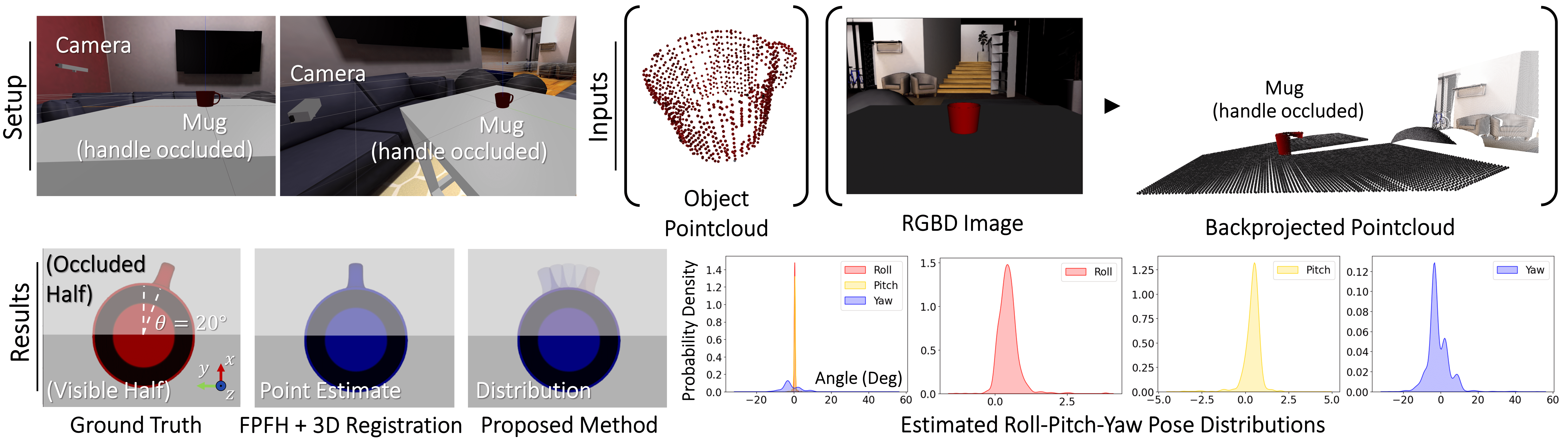}
  \caption{A proof of concept experiment that requires estimating the 6DOF pose of a textureless mug from a single RGBD image when its handle is occluded. As shown in the plots, the high uncertainty of the yaw angle (i.e., due to the mug being symmetric along the z axis) is appropriately captured.}
  \label{fig:fig1}
\end{figure*}

\subsection{Preliminary Definitions and Assumptions}
\textbf{Objects:} An object is represented as a structured pointcloud $\mathcal{O} = \{(\mathbf{p}_i, \mathbf{z}_i)\}_{i=1}^{N}$ (i.e., a pointcloud where every point $i$ stores a feature descriptor $\mathbf{z}_i \in \mathbb{R}^d$ in addition to its coordinates $\mathbf{p}_i \in \mathbb{R}^3$). In practice, such a structured pointcloud can be obtained from posed RGBD images by backprojecting the spatial features of a computer vision backbone \cite{caron2021emerging, radford2021learning}. Given a pose $(\mathbf{R}, \mathbf{t}) \in SO(3) \times \mathbb{R}^3$, $\mathcal{O}_{\mathbf{R}, \mathbf{t}} = \{(\mathbf{q}_i, \mathbf{z}_i)\}_{i=1}^{N}$ denotes the structured pointcloud resulting from rotating and translating $\mathcal{O}$ (i.e., $\mathbf{q}_i = \mathbf{R} \mathbf{p}_i + \mathbf{t}$).

\textbf{Scene:} A scene is represented as a function $f_{scn}: \mathbb{R}^3 \rightarrow \mathbb{R}^d$ that maps every 3D coordinate $\mathbf{x}$ to a feature descriptor $f_{scn}(\mathbf{x}) = \mathbf{z}$. In practice, $f_{scn}$ can be constructed by creating a structured pointcloud of a scene, and then fusing it into a dense representation (e.g., using kinect-fusion and extensions \cite{newcombe2011kinectfusion, jatavallabhula2023conceptfusion, curless1996volumetric}, or implicit neural representations \cite{simeonov2022neural, Peng2023OpenScene}). We assume that 3D coordinates that are observed to be empty space are mapped to a special descriptor $\mathbf{z}_{emp}$ by $f_{scn}$, while coordinates that are unobserved (e.g., due to occlusion) are mapped to another special descriptor $\mathbf{z}_{null}$.

\textbf{Object Classifier:} We further assume that for each object $\mathcal{O}$ to be localized, a binary classifier $p_{\mathcal{O}}: \mathbb{R}^3 \rightarrow [0, 1]$ is available that maps every 3D coordinate $\mathbf{x}$ to its probability $p_{\mathcal{O}}(\mathbf{x})$ of belonging to $\mathcal{O}$. In practice, $p_{\mathcal{O}}$ can be obtained from  RGBD images by backprojecting the classification scores of a semantic segmentation model \cite{he2017mask} or the similarity heatmaps produced by a classical template matching method \cite{dekel2015best}. This can be done concurrently with the construction of $f_{scn}$, as part of a dense SLAM pipeline \cite{rosinol2020kimera, dellaert2012factor, cadena2016past}.

\subsection{Probabilistic Localization of an Object}
\textbf{Visual Similarity Between Two Points:} We assume that the inner-product $\mathbf{z}_1^T\mathbf{z}_2$ in the feature descriptor space captures a notion of visual similarity, which is a common empirical observation for many foundational deep-learning models \cite{caron2021emerging, radford2021learning, oquab2023dinov2}.
We extend the definition of this inner-product such that $\mathbf{z}^T\mathbf{z}_{emp} = -\infty$ and $\mathbf{z}^T\mathbf{z}_{null} = 0$ hold for all $\mathbf{z}$.
Given two points $(\mathbf{p}_1, \mathbf{p}_2)$ and their corresponding feature descriptors $(\mathbf{z}_1, \mathbf{z}_2)$, we define the probability of $\mathbf{p}_1$ and $\mathbf{p}_2$ being \say{visually identical} as $\frac{1}{Z}e^{\, \beta \, \mathbf{z}_1^T\mathbf{z}_2}$, where $Z$ is a constant normalization factor\footnote{As is customary \cite{thrun2002probabilistic}, the constant normalization factor $Z$ is used as a catch-all notation for all unnormalized probability distributions, even if the actual value of $Z$ can differ across different distributions. Furthermore, since this constant factor has no influence on inference or sampling, it is discarded from most equations involving probabilities.} and $\beta$ is the temperature hyperparameter. 
We note that this is the same functional form commonly used in contrastive losses for self-supervised representation learning (e.g., InfoNCE \cite{oord2018representation}, SGP \cite{yang2021self}).

\textbf{Probabilistic Localization:} We formulate the localization problem by postulating a number of elementary probabilistic events (e.g., in relation to single points), and deriving the probability densities for more complex events (e.g., in relation to objects) from intersecting these elementary events. 
For any point $\mathbf{q}_i$ on $\mathcal{O}_{\mathbf{R}, \mathbf{t}}$, we define the event that \say{$\mathbf{q}_i$ is correctly localized} as $\mathbf{q}_i$ overlapping a scene point $\mathbf{x}$ which: i) belongs to object $\mathcal{O}$ (as classified by $p_{\mathcal{O}}$), \textit{and} ii) is visually identical to $\mathbf{z}_i$ (as captured by $f_{scn}(\mathbf{q}_i)^T \mathbf{z}_i)$). Given our previous definitions for these two events (and assuming independence), the probability density for their intersection would be derived as 
\begin{equation}\label{eq:0}
p_{loc}(\mathbf{q}_i) = \frac{1}{Z} \, p_{\mathcal{O}}(\mathbf{q}_i) \, e^{\, \beta \, f_{scn}(\mathbf{q}_i)^T \mathbf{z}_i} \quad . 
\end{equation}
It is worth emphasizing that instead of assigning a binary localization label (i.e., \say{correct} or \say{incorrect}) to a given coordinate hypothesis $\mathbf{q}_i$ (e.g., as in one-to-one point matching), this formulation instead assigns a more relaxed notion of a continuous localization probability.\footnote{
While not required by the proposed formulation, if assigning a binary decision "correct" or "incorrect" to $\mathbf{q}_i$ becomes necessary, one can threshold $p_{loc}(\mathbf{q}_i)$. Rather than making such a discrete decision for the correctness of a point localization, the proposed formulation instead opts for propagating the point localization uncertainties (i.e., represented by their associated densities $p_{loc}(\mathbf{q}_i)$) all the way to a final pose distribution $p(\mathcal{O}_{\mathbf{R}, \mathbf{t}})$.} Similarly, we define the event that \say{object $\mathcal{O}_{\mathbf{R}, \mathbf{t}}$ is correctly localized} as \textit{all} points $\mathbf{q}_i$ on it being correctly localized. The log probability for this event (again assuming independence) would be
\begin{equation}\label{eq:1}
\begin{split}
\log p(\mathcal{O}_{\mathbf{R}, \mathbf{t}}) &= \sum_{i=1}^{N} \log p_{loc}(\mathbf{q}_i) \\
&= \sum_{i=1}^{N} \log p_{\mathcal{O}}(\mathbf{q}_i) + \beta \sum_{i=1}^{N} f_{scn}(\mathbf{q}_i)^T \mathbf{z}_i \quad , 
\end{split}
\end{equation}
\noindent where $\mathbf{q}_i = \mathbf{R} \mathbf{p}_i + \mathbf{t}$ and the factor $\log Z$ was discarded. 

\textbf{Connections to two-stage pose estimation:} Inspecting \eqref{eq:1}, we see that its maximization practically implements a probabilistic relaxation of the traditional two-stage object pose estimation paradigm (i.e., correspondence detection, followed by geometric alignment). In particular, the term $\sum_{i=1}^{N} f_{scn}(\mathbf{q}_i)^T \mathbf{z}_i$ captures a soft notion of visual correspondence (i.e., maximize total visual similarity), while the term $\sum_{i=1}^{N} \log p_{\mathcal{O}}(\mathbf{q}_i)$ captures a soft notion of point-set registration (i.e., align points $\mathbf{q}_i$ from the object pointcloud to the points segmented by $p_{\mathcal{O}}$ in the scene pointcloud). 
\subsection{Estimating Most-Probable Pose Hypotheses}
Given our definition of $\log p(\mathcal{O}_{\mathbf{R}, \mathbf{t}})$, it is tempting to optimize \eqref{eq:1} with respect to the pose $(\mathbf{R}, \mathbf{t})$, which effectively corresponds to a maximum likelihood estimate (MLE). An important consideration is that the distribution $p(\mathcal{O}_{\mathbf{R}, \mathbf{t}})$ can be highly multimodal, particularly for objects and scenes that are textureless, symmetrical, and/or highly occluded where the correct pose is uncertain or there are multiple equally valid solutions such as Fig.\ref{fig:fig1}. For such cases, explicitly representing the entire distribution $p(\mathcal{O}_{\mathbf{R}, \mathbf{t}})$ has advantages over a point estimate, and can allow a downstream planning and control pipeline to reason about the resulting uncertainty (e.g., grasping the occluded handle of the textureless mug).
Therefore, we propose representing the entire distribution $p(\mathcal{O}_{\mathbf{R}, \mathbf{t}})$ by returning a set of highly probable samples drawn from it (i.e., a particle filter representation \cite{thrun2002probabilistic}). This involves two steps: i) maximizing the MLE objective \eqref{eq:1} with a deterministic continuous optimization routine to obtain a point estimate, ii) using this point estimate to initialize (i.e., for faster convergence) a black-box Monte Carlo routine for sampling from the unnormalized distribution $p(\mathcal{O}_{\mathbf{R}, \mathbf{t}})$.

\textbf{MLE for $p(\mathcal{O}_{\mathbf{R}, \mathbf{t}})$:} Directly optimizing the MLE objective \eqref{eq:1} with respect to $(\mathbf{R}, \mathbf{t})$ is difficult for two main reasons:
\begin{itemize}
    \item It is not differentiable due to those points where an inner product $\mathbf{z}^T\mathbf{z}_{emp} = -\infty$ needs to be evaluated for empty-space constraints.
    \item Due to terms $f_{scn}(\mathbf{R} \mathbf{p}_i + \mathbf{t})$ and $\log p_{\mathcal{O}}(\mathbf{R} \mathbf{p}_i + \mathbf{t})$, the resulting optimization problem is not of a form where an efficient and certifiably optimal global solver exists.
\end{itemize}

To address these difficulties, we first define a 3D grid of points $\{\mathbf{y}_j\}_{j=1}^{M}$, where $f_{scn}(\mathbf{y}_j)$ is different from $\mathbf{z}_{emp}$ and $\mathbf{z}_{null}$ for all $j$. We can then approximate the MLE objective \eqref{eq:1} with the following standard robust estimation form:
\begin{equation}\label{eq:2}
\min_{\mathbf{R}, \mathbf{t}} \ \sum_{i=1}^{N} \sum_{j=1}^{M} \, c_{ij} \, \rho(||\mathbf{y}_j - \mathbf{R} \mathbf{p}_i + \mathbf{t}|| \, , \, \beta_i) \quad , 
\end{equation}
\noindent where $c_{ij}= \log p_{\mathcal{O}}(\mathbf{y}_j) + \beta f_{scn}(\mathbf{y}_j)^T \mathbf{z}_i$ (i.e., the log-probability for correctly localizing the point $\mathbf{y}_j$ \eqref{eq:1}), and $\rho(r, \beta_i)$ is a robust cost function like truncated least-squares \cite{yang2022certifiably}. We note that minimizing \eqref{eq:2} requires aligning $\mathbf{p}_i$ to those $\mathbf{y}_j$ that would have the highest probability of being a correct localization for it, which therefore approximates the MLE objective \eqref{eq:1} (i.e., where the approximation quality depends on the resolution of the 3D grid $\{\mathbf{y}_j\}_{j=1}^{M}$). There are two important things to note about \eqref{eq:2}: i) efficient and certifiably optimal global solvers for it exist \cite{yang2020graduated, yang2020teaser} which can be used off-the-shelf, and ii) it can be converted into a polynomial optimization problem (i.e., by applying the Black-Rangarajan
duality \cite{black1996unification} to convert $\rho(r, \beta_i)$), which makes the associated mathematical machinery readily applicable (e.g., solution through a sparse semidefinite relaxation routine like STRIDE \cite{yang2022certifiably} for polynomial optimization).

\textbf{Sampling Poses from $p(\mathcal{O}_{\mathbf{R}, \mathbf{t}})$:} Equation \eqref{eq:1} provides a way to compute an unnormalized probability density $p(\mathcal{O}_{\mathbf{R}, \mathbf{t}})$ for any pose hypothesis $(\mathbf{R}, \mathbf{t})$. 
In statistics, obtaining samples from a given unnormalized probability density function (i.e., called the sampling problem \cite{gelman2013bayesian}) is typically solved using Markov chain Monte Carlo (MCMC) methods \cite{brooks2011handbook}. Therefore, off-the-shelf MCMC algorithms can be utilized to generate particles (i.e., object pose hypotheses $(\mathbf{R}, \mathbf{t})$) from $p(\mathcal{O}_{\mathbf{R}, \mathbf{t}})$.
Given the close theoretical connection between sampling and global optimization \cite{ma2019sampling}, more general stochastic optimization algorithms like differential evolution \cite{storn1997differential, price2013differential} or simulated annealing \cite{kirkpatrick1983optimization, bertsimas1993simulated} can also be employed.
The convergence properties of such Monte Carlo routines are highly dependent on their initialization, and the MLE routine \eqref{eq:2} is particularly suitable for this purpose.

\section{Implementation and Conclusion}
\textbf{Proof of Concept:} We implement a demonstration of the proposed framework within the Drake simulation environment \cite{drake}. As shown in Fig.\ref{fig:fig1}, the task is to estimate the 6DOF pose of a textureless symmetric mug from an RGBD image when its handle is occluded, and the proposed pipeline can successfully recover a feasible distribution for all likely poses. For more details, please refer to the appendix.

\textbf{Conclusion and Future Work:} Potential advantages of the proposed formulation over existing object pose estimation pipelines \cite{yang2023object} are that: i) it doesn't require committing to any given definition of keypoints (i.e., or associated methods for their detection and matching), and ii) it explicitly reasons about empty space constraints (i.e., $\mathbf{z}^T\mathbf{z}_{emp} = -\infty$), which are known to provide a very rich signal for geometric reasoning \cite{park2019deepsdf}. While this preliminary letter focuses on presenting the mathematical underpinnings of the proposed framework, a large scale quantitative evaluation on well established benchmarks is the most immediate direction for future work. Another important direction is to speed-up the Monte Carlo routine employed for sampling from $p(\mathcal{O}_{\mathbf{R}, \mathbf{t}})$, which currently prevents real-time implementation. Promising approaches include developing custom importance samplers tailored to \eqref{eq:1} (e.g., by injecting noise into $c_{ij}$ in \eqref{eq:2} and repeatedly solving the resulting MLE problem for different realizations of noise) rather than using black box Monte Carlo methods off-the-shelf, or entirely discarding Monte Carlo approaches and switching to Bayesian optimization methods \cite{hennig2022probabilistic, hennig2012entropy}. A more general direction for future work is to design planning and control algorithms that can make effective use of the uncertainty encoded in the entire distribution $p(\mathcal{O}_{\mathbf{R}, \mathbf{t}})$, for example by using $p(\mathcal{O}_{\mathbf{R}, \mathbf{t}})$ to define potential functions for geometric control \cite{van2022geometric, ratliff2021generalized, ratliff2020optimization}, or using $p(\mathcal{O}_{\mathbf{R}, \mathbf{t}})$ to maximize information gain \cite{thrun2002probabilistic} for efficient trial-and-error learning (e.g., for robotic grasping).

\section*{APPENDIX}
We present a detailed step-by-step explanation of the pipeline implemented in Fig\ref{fig:fig1} for clarity:

\textbf{Preparing the inputs:} The inputs to the pipeline are a single RGBD image $I_{scn}$ of the apartment scene (together with camera intrinsics), and four posed RGBD images $\{I_n\}_{n=1}^4$ of the mug captured in isolation at a location different from the apartment scene. All images have a resolution of $640\times480$. To construct $f_{scn}$, we first do a forward pass with $I_{scn}$ using the DINO transformer backbone \cite{caron2021emerging}, take the $14\times14$ spatial tokens of the final transformer layer, and upsample this $14\times14$ grid of tokens (where each spatial token is a 384 dimentional floating point vector) to $640\times480$ resolution through bilinear interpolation to form the feature descriptors $\{z_i\}_{scn}$. These descriptors are then backprojected into a structured pointcloud \cite{KaolinLibrary} using the depth values and camera intrinsics, and uniformly voxel-downsampled by assigning every voxel the mean of the feature descriptors of all points within it. The values $f_{scn}(x)$ for continuous coordinates are constructed by bilinear interpolation of the feature descriptors of four neighboring voxels to $x$. To construct the structured pointcloud $\mathcal{O}$ for the mug, we again do a forward pass with $\{I_n\}_{n=1}^4$ using DINO, upsample the resulting spatial tokens, segment the mug in each image using GrabCut \cite{rother2004grabcut}, and then backproject feature descriptors from pixels corresponding to the mug in all four images into a pointcloud. To construct the classifier $p_{\mathcal{O}}$, we compute the best-buddy similarity score \cite{dekel2015best} between the feature descriptors of every point in $\mathcal{O}$ and every voxel in $f_{scn}$, and threshold the result to give a noisy binary classification output. For negative points, $\log p_{\mathcal{O}}$ is set to a large negative value $c_{min}$ rather than $-\infty$.

\textbf{Probabilistic Localization:} Given $f_{scn}$, $\log p_{\mathcal{O}}$, and $\mathcal{O}$ constructed in the previous section, evaluating $\log p(\mathcal{O}_{\mathbf{R}, \mathbf{t}})$ in \eqref{eq:1} for any pose $(\mathbf{R}, \mathbf{t})$ requires a vectorized sum across all points in $\mathcal{O}$, which takes $\sim 2e^{-3}$ seconds in total. To generate particles (i.e., pose hypotheses) from $\log p(\mathcal{O}_{\mathbf{R}, \mathbf{t}})$, we run a stochastic global optimization algorithm called two-point differential evolution \cite{storn1997differential}, using the highly efficient implementation provided by Nevergrad \cite{nevergrad}. The algorithm takes $\sim 10$ seconds to generate $1e^4$ particles. As is customary with particle filtering \cite{thrun2002probabilistic}, the particles are then assigned importance weights using the unnormalized distribution $p(\mathcal{O}_{\mathbf{R}, \mathbf{t}})$, and resampled with replacement using these weights to approximate $1e^4$ samples drawn from $p(\mathcal{O}_{\mathbf{R}, \mathbf{t}})$. We also experimented with directly sampling from $p(\mathcal{O}_{\mathbf{R}, \mathbf{t}})$ using Markov chain Monte Carlo with PyMC \cite{salvatier2016probabilistic}, which didn't converge within five minutes of wall-clock time. As our approach converges sufficiently fast from a random initialization, we didn't need to use MLE for initialization. To generate continuous distributions (e.g., Fig.\ref{fig:fig1}) using the $1e^4$ pose hypotheses sampled from $p(\mathcal{O}_{\mathbf{R}, \mathbf{t}})$, we employ gaussian kernel density estimation as implemented in SciPy \cite{2020SciPy-NMeth}.

\textbf{Validity of the Proposed Probabilistic Models:} The derivations for the probability densities $p_{loc}(\mathbf{q}_i)$ and $p(\mathcal{O}_{\mathbf{R}, \mathbf{t}})$ involve a number of assumptions such as the inner-product $\mathbf{z}_1^T\mathbf{z}_2$ capturing visual similarity, the functional form $\frac{1}{Z}e^{\, \beta \, \mathbf{z}_1^T\mathbf{z}_2}$ and the classifier $p_{\mathcal{O}}(\mathbf{x})$ being well-calibrated distributions, as well as independence assumptions. Therefore it is important to note that the resulting probability densities can at best be described as pragmatic models of uncertainty whose motivations and merits are mainly practical (e.g, robustifying downstream planning and control).

\bibliographystyle{IEEEtran} 
\bibliography{root}

\end{document}